\documentclass[10pt,twocolumn,letterpaper]{article}

\usepackage[pagenumbers]{iccv} 

\usepackage{pifont}
\usepackage{subcaption} 
\usepackage{xspace}
\usepackage{dsfont}
\usepackage{array}
\usepackage{colortbl}
\usepackage{xcolor}
\usepackage{algorithm}
\usepackage{algpseudocode}
\usepackage{listings}  
\usepackage{graphicx}  

\definecolor{darkgreen}{RGB}{0,128,0}  

\newcommand{\ours}{OV-SCAN\xspace}
\newcommand{\NOD}{SC-NOD\xspace}

\newcommand{\smallskipcustom}{}

\definecolor{cvprblue}{rgb}{0.21,0.49,0.74}
\usepackage[pagebackref,breaklinks,colorlinks,allcolors=cvprblue]{hyperref}

\def\paperID{9080} 
\def\confName{ICCV}
\def\confYear{2025}

\title{\ours: Semantically Consistent Alignment for Novel Object Discovery in Open-Vocabulary 3D Object Detection}

\author{Adrian Chow \quad Evelien Riddell \quad Yimu Wang \quad Sean Sedwards \quad Krzysztof Czarnecki\\
University of Waterloo\\
}

\begin{document}
\maketitle
\begin{abstract}
Open-vocabulary 3D object detection for autonomous driving aims to detect novel objects beyond the predefined training label sets in point cloud scenes. 
Existing approaches achieve this by connecting traditional 3D object detectors with vision-language models (VLMs) to regress 3D bounding boxes for novel objects and perform open-vocabulary classification through cross-modal alignment between 3D and 2D features. However, achieving robust cross-modal alignment remains a challenge due to semantic inconsistencies when generating corresponding 3D and 2D feature pairs. To overcome this challenge, we present OV-SCAN, an Open-Vocabulary 3D framework that enforces Semantically Consistent Alignment for Novel object discovery. \ours employs two core strategies: discovering precise 3D annotations and filtering out low-quality or corrupted alignment pairs (arising from 3D annotation, occlusion-induced, or resolution-induced noise). Extensive experiments on the nuScenes dataset demonstrate that \ours achieves state-of-the-art performance. 

\end{abstract}

\section{Introduction} \label{sec:intro}

Recent advances in autonomous driving have intensified research in 3D object detection, with most methods operating in a closed-set setting, classifying objects into a small set of predefined categories (e.g. vehicle, cyclist, pedestrian). However, such high-level object identification is inadequate for navigating complex real-world environments. Autonomous systems need to recognize objects at a deeper semantic level. For instance, distinguishing between a fire truck and a school bus, rather than simply labeling them as trucks or buses, is critical for safe and context-aware decision-making. 

To address the need for deeper semantic understanding in 3D vision systems, open-vocabulary (OV)-3D object detection has emerged. In OV-2D object detection, OV capabilities rely on large-scale datasets of 
image-text pairs, which are often unavailable in 3D~\cite{liu2024DINO, li_grounded_2022, minderer_scaling_2023, cheng_yolo-world_2024, yao_detclip_2022}. Without similar large-scale 3D-text datasets, methods must find 
\begin{figure}[h!]
    \centering
    \includegraphics[width=0.80\linewidth]{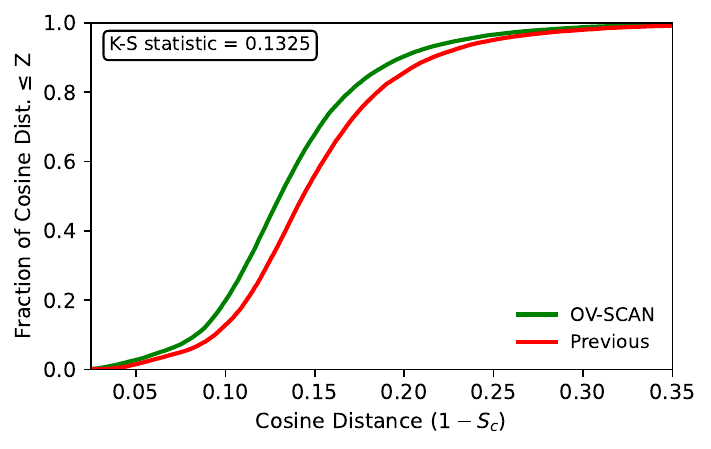}
    \vspace{-1em}
    \caption{
     \textbf{Cross-modal Alignment Performance.} The red CDF shows the distribution of the distance between the 3D embedding produced by a baseline OV-3D detector and the corresponding 2D embedding from CLIP on the nuScenes validation set. The green CDF shows this distribution using OV-SCAN instead of the baseline. The latter is shifted to the left, showing an improved alignment between the 3D and 2D embeddings due to supervision by our higher quality 3D-2D proposal pairings and our alignment head.
    }
    \vspace{-1em}
    \label{fig1}
\end{figure}
alternative ways to provide accurate 3D annotations for novel objects and achieve alignment between 3D and text.

OV-3D object detection faces two main challenges: (1) novel object discovery (NOD), which involves generating 3D labels for novel objects in order to train an off-the-shelf LiDAR-based detector, and (2) semantic alignment between 3D features and textual embeddings. Existing NOD approaches generally fall into two categories, online and offline methods. The former starts with a set of base objects and progressively expand the training labels to include novel objects during training ~\cite{cao_coda_2023, cao2024collaborativenovelobjectdiscovery}. The latter generates pseudo-labels for novel objects ahead of time ~\cite{lu_open-vocabulary_2023, leonardis_find_2025, zhang_opensight_2024}. Online methods often struggle to balance training with discovery, leading to biases in object shape and scale that result in overfitting to base categories. Due to the aforementioned challenges with online methods, recent work has increasingly shifted toward offline approaches. In these offline pipelines, OV-2D detectors such as Grounding DINO~\cite{liu2024DINO} and OWL-ViT~\cite{minderer2022Owl} first generate 2D proposals from multi-view images, which are typically paired with point cloud. These 2D proposals are then projected into 3D space and reformulated as 3D pseudo-labels for training. However, because LiDAR point clouds are inherently sparse, naïve projection methods such as simple clustering struggle to accurately fit boxes to objects that are only partially observable and not visibly dense. Most recently, Find n' Propagate~\cite{leonardis_find_2025} employs an exhaustive greedy search to optimally choose 3D bounding box parameters for each 2D proposal, yet this approach still suffers from several 3D annotation errors shown in \cref{fig2}. Such noise hinders the 3D object detector’s ability to extract discriminative 3D features.

To achieve semantic alignment between 3D data and text, existing methods use the 2D image space as an implicit intermediary~\cite{zhang_opensight_2024, lu_open-vocabulary_2023, cao_coda_2023, zhang_fm-ov3d_2024}. Each 3D bounding box label is associated with a corresponding 2D image region from which a visual embedding is extracted using a VLM such as CLIP~\cite{radford2021CLIP}. During training, a base 3D object detector is supervised using the 3D annotations for box regression, while simultaneously aligning 3D object features with target 2D image embeddings. During test time, novel objects are classified via prompt-based classification, assigning labels based on the highest similarity between 3D features and text embeddings. Achieving strong alignment requires both semantically distinct 3D and 2D features. However, existing methods often overlook common autonomous driving scenarios where objects are partially occluded (\cref{fig:occlusion-noise-example}) or appear small at longer distances (\cref{fig:resolution-noise-example}). In such cases, the 2D features become ambiguous or lack sufficient representation, leading to confusion during cross-modal alignment.

In this work, we introduce \ours, a OV-3D object detector which targets strong OV detection performance from enhanced semantic alignment between 3D objects and 2D image features. More specifically, we introduce the Semantically-Consistent Novel-Object Discovery (\NOD) module  to handle the inherit challenges of noisy cross-modal alignment. To address \textit{3D annotation noise}, \NOD reformulates 3D box search as a non-linear optimization problem and employs an adaptive search strategy to efficiently optimize 3D bounding box parameters from 2D proposals. To mitigate \textit{occlusion-induced noise} and \textit{resolution-induced noise}, SC-NOD enables a selective alignment mechanism by identifying the alignment pairs that are semantically consistent. During training, all 3D annotations contribute to box-related losses in order to maintain high recall, while only semantically consistent alignment pairs guide the cross-modal alignment. We further propose the Hierarchical Two-Stage Alignment (H2SA) head to enhance cross-modal alignment. Novel classes are organized hierarchically by first grouping them into broad, high-level categories (e.g., car) and then subdividing these categories into more detailed subclasses (e.g., SUV, sedan). This hierarchical structure allows H2SA to initially perform coarse-grained classification, before carefully aligning class-informed object features for fine-grained discrimination among closely related subclasses. We summarize our main contributions as followed:
 
\begin{figure}
    \centering
    \includegraphics[width=\linewidth]{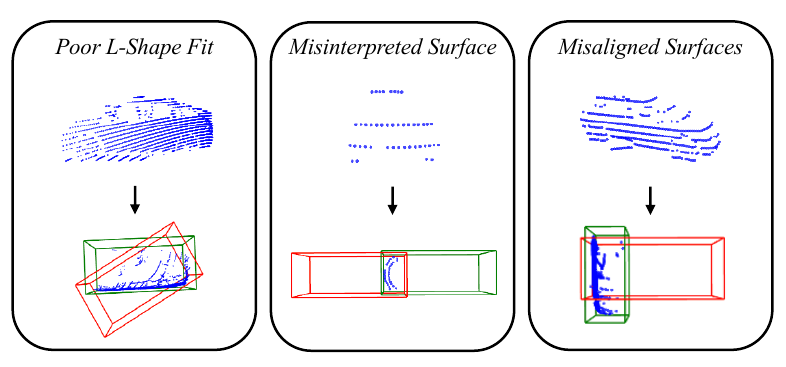}
    \caption{\textbf{3D Annotation Errors.} Common 3D annotation errors during box parametrization, including but not limited to, poor L-shape fitting, misinterpreted surfaces, and misaligned surfaces.}
    \label{fig2}
\end{figure}
\begin{figure}
    \centering
    \begin{subfigure}[b]{0.49\linewidth}
        \centering
        \includegraphics[width=\linewidth]{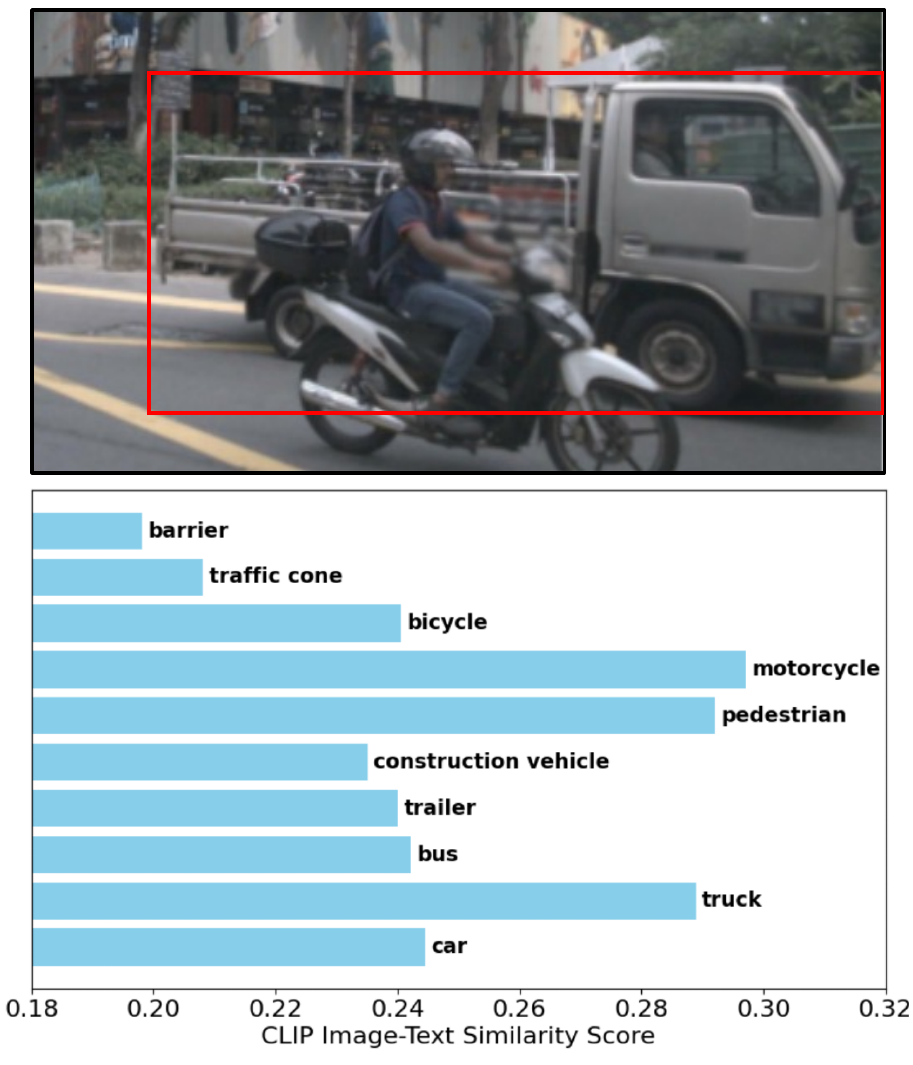}
        \caption{Occlusion-Induced Noise}
        \label{fig:occlusion-noise-example}
    \end{subfigure}
    \hfill
    \begin{subfigure}[b]{0.49\linewidth}
        \centering
        \includegraphics[width=\linewidth]{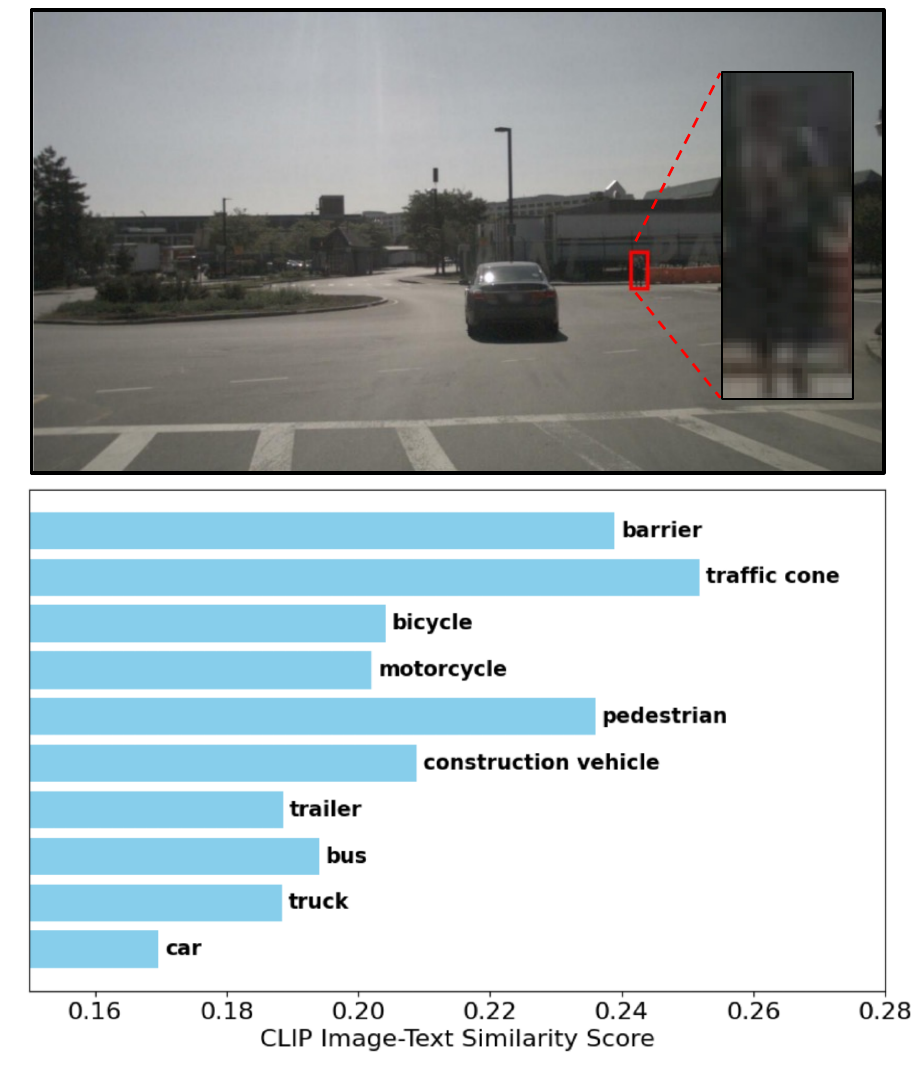}
        \caption{Resolution-Induced Noise}
        \label{fig:resolution-noise-example}
    \end{subfigure}
    \caption{\textbf{Sources of Semantic Discrepancies.} (a) CLIP similarity scores for a truck reveals that occlusion cases result in an ambiguous 2D image feature.
    (b) CLIP similarity scores for a distant pedestrian demonstrate that insufficient resolution leads to degraded 2D image feature.}
    \label{fig:main}
    \vspace{-3.75mm}
\end{figure}

\begin{itemize}
    \item We present \ours, an OV-3D object detector benefiting from improved cross-modal alignment, see \cref{fig1}.
    \item We propose the \NOD module, extending existing 3D box search methods and introducing selective cross-modal alignment during training. In addition, the proposed H2SA head effectively aligns 3D-to-2D alignment pairs by introducing a two-stage alignment process.
    \item We validate \ours on the nuScenes~\cite{caesar2020nuscenes} and KITTI~\cite{geiger2013kitti} datasets, demonstrating that \ours achieves state-of-the-art performance in an OV-3D setting without supervising with any human-annotated data.
\end{itemize}

\section{Related Work}

\subsection{Traditional 3D Object Detection}

Traditional 3D object detection methods focus primarily on closed-set scenarios, training models on a limited number of common classes. LiDAR-based approaches are categorized into single-representation, hybrid-representation, and transformer-based methods. Single-representation methods such as VoxelNet~\cite{zhou_voxelnet_2018} utilize voxel grids processed by 3D CNNs but are computationally intensive. Improvements include sparse convolutions~\cite{yan_second_2018, chen_voxelnext_2023}, pillars~\cite{lang_pointpillars_2019}, and BEV representations~\cite{yin_center-based_2021} for efficiency. Point-based methods like PointNet~\cite{qi_pointnet_2017} and PointRCNN~\cite{shi_pointrcnn_2019} directly process raw point clouds, achieving detailed geometry at the cost of speed. Hybrid models like PV-RCNN~\cite{shi_pv-rcnn_2020, shi_pv-rcnn_2023} integrate voxel and point features for improved accuracy. Transformer-based architectures~\cite{sun_swformer_2022, ning_dvst_2023, misra_end--end_2021, liu_flatformer_2023, mao_voxel_2021} deploy self-attention to capture global context. LiDAR-camera fusion approaches~\cite{bai_transfusion_2022, liu_bevfusion_2023, yin_is-fusion_2024} combine LiDAR’s spatial precision with RGB images to enhance semantic understanding.

\subsection{Open-Vocabulary 2D Object Detection}

The emergence of large-scale pretrained models with strong text-vision alignment and impressive zero-shot transfer capabilities has driven significant interest in tackling the OV problem in 2D object detection. Approaches like OVR-CNN~\cite{zareian_open-vocabulary_2021}, ViLD~\cite{gu_open-vocabulary_2022}, OWL-ViT~\cite{minderer2022Owl}, and OV-DETR~\cite{zang_open-vocabulary_2022} address this challenge through knowledge distillation, using pretrained VLMs such as CLIP~\cite{radford2021CLIP}. These methods integrate semantic knowledge into traditional detectors by contrastively aligning region features or object embeddings. Subsequently, GLIP~\cite{li_grounded_2022} reformulates object detection as phrase grounding, exploiting large-scale image-text pair pretraining to unify localization and classification while enabling generalization to unseen categories through semantic understanding. Grounding DINO~\cite{liu2024DINO} advances this idea further by applying tightly coupled fusion strategies on top of DINO~\cite{caron_emerging_2021}, ensuring effective integration of language and vision. Extending OV-2D object detection to real-time applications, YOLO-World~\cite{cheng_yolo-world_2024} uses vision-language modelling within the efficient YOLO architecture~\cite{Jocher2023YOLOv8}.

\subsection{Open-Vocabulary 3D Object Detection}

Building on the success of OV-2D object detectors, recent efforts have aimed to extend these capabilities to 3D. A key challenge in this adaptation is the scarcity of diverse labeled data required for large-scale 3D training. OV-3Det~\cite{lu_open-vocabulary_2023} introduced the 2D-to-3D paradigm in an indoor setting, where VLMs generate annotations for a diverse set of objects, which are then lifted to 3D using simple point cloud clustering. CoDA~\cite{cao_coda_2023} pretrains a base detector with a small set of human-annotated labels and introduces a framework to incorporate novel labels during training. FM-OV3D~\cite{zhang_fm-ov3d_2024} further improves detection performance by exploiting the strengths of multiple foundational models. ImOpenSight~\cite{zhang_opensight_2024} extends open-vocabulary detection into the outdoor domain, incorporating temporal and spatial awareness in box annotations. In addition, their method also introduces 2D-to-3D geometric priors to guide annotation under sparse point cloud conditions. Find n' Propagate~\cite{leonardis_find_2025} addresses the issue of insufficient recall of novel objects and propagates detection capabilities to more distant areas. ImOV3D ~\cite{yang2024imov3d} demonstrates an effective method to train a OV-3D object detector without any 3D point clouds, instead using a pseudo-multimodal representation. Despite advancements, existing methods often neglect the impact of noise sources that weaken cross-modal alignment.
\begin{figure*}[!ht]
    \centering
    \includegraphics[width=1.0\textwidth]{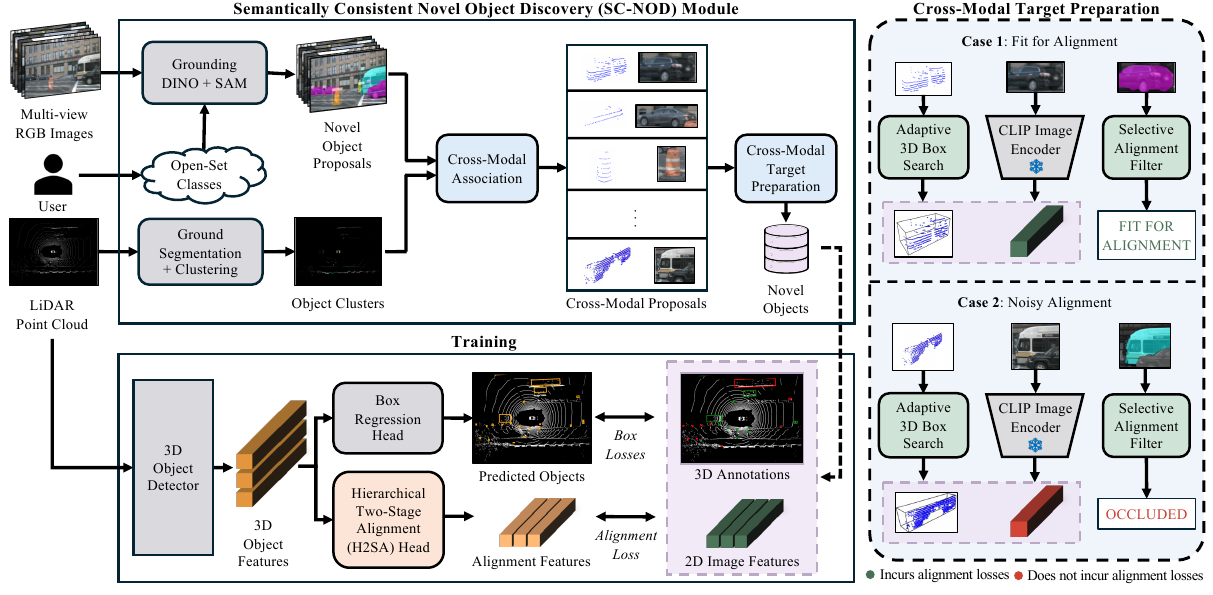}
    \caption{\textbf{Overall Framework for \ours.} During novel object discovery, \NOD associates novel object proposals with corresponding object clusters, creating cross-modal proposals. \NOD performs an adaptive search to fit 3D annotations and extracts 2D image features to prepare cross-modal targets for supervision. \NOD identifies which samples are fit for alignment based on cross-modal semantic consistency. During training, all 3D annotations are used, while only consistent novel objects guide cross-modal alignment.
    }
    \label{fig:3}
\end{figure*}

\section{Method} \label{method}
In this section, we detail \ours. An overview is provided in \cref{fig:3}. 
\NOD (\cref{sec:scnod}) generates 3D annotations from 2D proposals to form cross-modal alignment pairs. \NOD then determines the subset of semantically consistent pairs to guide cross-modal alignment. \ours comprises an off-the shelf 3D object detector with a specialized alignment head (\cref{sec:model-architecture}), trained using the novel objects detected by \NOD (\cref{sec:training}).

\subsection{Notation and Preliminaries}

\textbf{3D Object Detection.} In traditional LiDAR-based 3D object detection, the objective is to train a detector using input-target pairs \( \mathbf{D} = \{\mathcal{P}, \Omega\} \). Given a point cloud \(\mathcal{P} = \{p_i\}_{i=1}^M\) with \(M\) points, where each point \( p_i = (x_i, y_i, z_i) \) represents spatial coordinates, the detector predicts targets \(\Omega = \{(B_i, c_i)\}_{i=1}^N\) for \(N\) objects. Each target consists of a 3D bounding box \( B_i = (x_i, y_i, z_i, l_i, w_i, h_i, (r_y)_i) \) with center \((x_i, y_i, z_i)\), dimensions \((l_i, w_i, h_i)\), orientation \((r_y)_i\), and a class label \( c_i \). We assume that each LiDAR frame is accompanied by a set of \( K \) images from different perspectives (multi-view), represented as \( \mathcal{I} = \{ I_j \}_{j=1}^K \). LiDAR points can be mapped to image space by using the extrinsic and intrinsic calibrations \( T_\text{ext} \) and \( T_\text{int} \).

\smallskipcustom

\noindent \textbf{Open-Vocabulary 3D Object Detection.} In an OV setting, 3D object detectors do not have access any target labels \( \Omega \) and instead rely on VLMs to generate annotations for novel objects, creating a \emph{novel object bank}. Our method extends the traditional target pair of 3D bounding box and class label, into a triplet target denoted by \( {\Omega}' = \{(B_i, c_i, \mathcal{A}_{\text{2D},i})\}_{i=1}^N \).  \( \mathcal{A}_{\text{2D},i} \in \mathbb{R}^{D} \) represents a 2D alignment embedding that captures the semantic correspondence between the 3D object and its 2D image projection.  Traditional LiDAR-based 3D object detection methods are designed to regress 3D object features \( \mathcal{O}_\text{3D}  \in \mathbb{R}^{H} \) given the input point-cloud \(\mathcal{P}\). Our method aims to learn a function \( f \) that maps a set of 3D object features \( \mathcal{O}_\text{3D}\) to both the set of predicted 3D bounding boxes \( B \) and predicted alignment vectors \( \mathcal{A}_\text{2D} \), thereby bridging the 3D and 2D domains in OV object detection. These alignment features are then used for prompt-based classification by comparing them with text embeddings generated from class prompts, enabling fine-grained recognition of novel objects.

\subsection{Semantically Consistent NOD (\NOD)}
\label{sec:scnod}

\noindent \textbf{Novel Object Proposals}. A key component of open-vocabulary detection is identifying novel objects from an arbitrary set of classes. Following prior approaches~\cite{zhang_fm-ov3d_2024, zhang_opensight_2024, leonardis_find_2025}, we employ Grounding DINO~\cite{liu2024DINO} to detect proposals from a given list of novel classes. For each LiDAR frame, novel object proposals \( \mathbf{P}_{\text{2D}} \) are generated on \( K \) multi-view images, capturing objects from multiple perspectives. These proposals, defined as 2D bounding boxes, are represented for each image \( I_k \) as \( \mathbf{P}_{\text{2D},k} = \{ b_{k,j} \}_{j=1}^{N_k} \), where \( b_{k,j} \) specifies position and size. The complete set for a given frame is \( \mathbf{P}_{\text{2D}} = \bigcup_{k=1}^K \mathbf{P}_{\text{2D},k} \). SAM~\cite{kirillov_segment_2023} then pairs instance masks \( \mathbf{M} = \{ m_{k,j} \}_{k=1, j=1}^{K, N_k} \) with each proposal, where \( m_{k,j} \in \{0,1\}^{H \times W} \) denotes a binary mask for each \( b_{k,j} \), with \( m_{k,j}(x, y) = 1 \) indicating instance pixels. 

\smallskipcustom


\noindent \textbf{Cross-Modal Association}. During cross-modal association,  novel object proposals are paired with object clusters to form cross-modal proposals. To generate object clusters, we first deploy Patchwork++~\cite{lee_patchwork_2022} to segment and remove ground points. HDBScan~\cite{campello_density-based_2013} then clusters the remaining points into object clusters. To associate 2D bounding boxes with these 3D clusters, each 2D proposal is back-projected into a corresponding 3D frustum. Using the center frustum ray, 2D proposals are paired with object clusters exhibiting strong spatial correspondence. The pairs form the set of cross-modal proposals. Additional details and intricacies regarding this procedure is available in ~\cref{sec:cma_supp} of the supplementary.

\smallskipcustom

\noindent \textbf{Adaptive 3D Box Search}. For each cross-modal proposal, an amodal 3D bounding box is generated to accurately capture the object’s position, dimensions, and orientation in 3D. To mitigate the impact of 3D annotation noise, it is crucial to explore the parameter space effectively and identify the optimal configuration for novel objects. Find n’ Propagate~\cite{leonardis_find_2025} employs an exhaustive greedy search over a frustum-defined area to estimate box parameters for novel object proposals. However, this approach inefficiently distributes a large number of search candidates to suboptimal configurations due to its expansive search space and limited precision. Our method addresses this inefficiency by reformulating the task as a continuous nonlinear optimization problem within a localized search area. Additionally, while their cost function considers only point density and multi-view alignment, our method extends it to incorporate surface alignment, further mitigating annotation errors.

Given a cross-modal proposal, comprising a 2D bounding box \( b_{\text{img}} \) and a set of object points \( \mathcal{P}_{\text{obj}} \), the objective is to determine the 3D bounding box parameters \( \theta = (x, y, z, l, w, h, r_y) \) that best represent the object. Following similar works~\cite{leonardis_find_2025, zhang_opensight_2024, huang2024training}, the search is guided by constraining the bounding box dimensions to a predefined set of box anchors \( \mathbf{A} = \{(\mathbf{A}_{\text{min}}^i, \mathbf{A}_{\text{max}}^i)\}_{i=1}^C \), corresponding to each novel class. Our method relies on CLIP to classify the object into its corresponding novel class \( c \). To reduce ambiguity in orientation, the yaw angle \( r_y \) is constrained to the range \([0, \pi]\). The continuous nonlinear optimization problem is then formulated in standard form:
\begin{align}
\min_{\mathbf{\theta}} \, \mathcal{J}(\mathbf{\theta}, \mathcal{P}_{\text{obj}}, \mathbf{e}, b_{\text{img}}) 
= \; & \mathcal{J}_{\text{3D}}(\mathbf{\theta}, \mathcal{P}_{\text{obj}}, \mathbf{e}) 
+ \mathcal{J}_{\text{2D}}(\mathbf{\theta},b_{\text{img}})\, \nonumber \\
& \hspace{-4em} \text{subject to } \mathbf{A}_{\text{min}} \leq (l, w, h) \leq \mathbf{A}_{\text{max}} \, \nonumber \\
& \hspace{0.35em}0 \leq r_y \leq \pi, \
\label{eq:opt}
\end{align}
where \( \mathcal{J}_{\text{3D}} \) and $\mathcal{J}_{\text{2D}}$ are the cost functions detailed below, \( b_{\text{img}} \) is the 2D bounding box coordinates, \( \mathcal{P}_{\text{obj}} \) represents object points in 3D space, and \( \mathbf{e} \) is the ego's position in 3D space.

To evaluate the fit of the box with respect to the object points, \(\mathcal{J}_{\text{3D}}\) is defined as:
\begin{align}
\mathcal{J}_{\text{3D}}(\mathbf{\theta}, \mathcal{P}_{\text{obj}}, \mathbf{e}) = \, & \lambda_1 \mathcal{J}_{\text{density}}(\mathbf{\theta}, \mathcal{P}_{\text{obj}}) \nonumber \\
 + & \lambda_2 \mathcal{J}_{\ell\text{-shape}}(\mathbf{\theta}, \mathcal{P}_{\text{obj}}) \nonumber \\
 + & \lambda_3 \mathcal{J}_{\text{surface}}(\mathbf{\theta}, \mathbf{e})\,.
\end{align}

The point density term \(\mathcal{J}_{\text{density}}\) quantifies the proportion of the object’s points that are enclosed within the 3D box. This encourages the parameters to adjust so that all points are captured by the box. \(\mathcal{J}_{\text{density}}\) is defined as:
\begin{align}
\mathcal{J}_{\text{density}}(\theta, \mathcal{P}_{\text{obj}}) 
&= -\frac{\mathcal{G}_{\text{inside}}(\theta, \mathcal{P}_{\text{obj}})}{|\mathcal{P}_{\text{obj}}|},\label{eq:fdensity}
\end{align}
\noindent where
\begin{align}
 \mathcal{G}_{\text{inside}}(\theta, \mathcal{P}_{\text{obj}}) &= \sum_{p_i \in \mathcal{P}_{\text{obj}}} \mathds{1}\big(p_i \in B_{\text{3D}}(\theta)\big). 
\end{align}

The \(\ell\)-shape fitting term \(\mathcal{J}_{\ell\text{-shape}}\) measures how well points within the 3D bounding box align with two anchoring 2D edges, \(E_1\) and \(E_2\). Of the eight corners defining the box, the top four corners form edges, and the two closest to the ego reference point are selected. Minimizing the distance of object points \(p_i\) to the closest edge \(E \in \{E_1, E_2\}\)  ensures the bounding box aligns with the most relevant structural edges. In this case, \(d(p_i, E)\) is the euclidean distance from \(p_i\) to \(E\).
\(\mathcal{J}_{\ell\text{-shape}}\) is defined as:
\begin{multline}
\mathcal{J}_{\ell\text{-shape}}(\theta, \mathcal{P}_{\text{obj}})
= \\\sum\limits_{p_i \in \mathcal{P}_{\text{obj}}} \frac{\min\limits_{E \in \{E_1, E_2\}} d(p_i, E) \cdot  \mathds{1}(p_i \in B_{\text{3D}}(\theta)) }{\mathcal{G}_{\text{inside}}(\theta, \mathcal{P}_{\text{obj}})}.
\end{multline}

The surface fitting term \(\mathcal{J}_{\text{surface}}\) assigns object points to the box faces nearest to the ego, aligning surfaces closer to the ego reference point \(\mathbf{e}\). This is achieved by introducing a bias that pushes the bounding box away from the ego, clipped by a constant \( C_{\text{surface}} \) in the cost function. \(\mathcal{J}_{\text{surface}}\) is defined as:
\begin{align}
\!\mathcal{J}_{\text{surface}}(\theta, \mathbf{e}) \! 
&= \!-\!\min\left(\| (\theta_x, \theta_y) \! - \! (\mathbf{e}_x, \mathbf{e}_y) \|_2, C_{\text{surface}}\right)\!.\label{eq:f_ego}
\end{align}

Finally, consistency within the image space is measured by applying calibration transformations \( T_{\text{ext}} \) and \( T_{\text{int}} \). These transformations project the 3D bounding box \( \theta \) into its 2D counterpart in the image space using \( P_{\text{3D} \rightarrow \text{2D}} \). The resulting projection is evaluated against the original 2D bounding box using the IoU metric:
\begin{align}
\mathcal{J}_{\text{2D}}(\mathbf{\theta}, b_{\text{img}})
= \gamma \cdot \text{IoU}_\text{2D}(P_{\text{3D} \rightarrow \text{2D}}(\theta), b_{\text{img}}). \label{eq:mv_align}
\end{align}

The optimization is governed by a cost function that balances multiple objectives. The objectives are weighted by control constants (\(\lambda_1, \lambda_2,  \lambda_3, \gamma \)). 


Our method employs an evolutionary algorithm to iteratively refine the parameter space with increasing granularity. Box candidates are initialized within localized regions around object clusters to enhance exploration efficiency. Each scene frame is annotated, and results are compiled into a novel object bank. Duplicate 3D bounding boxes, arising from multi-view overlaps or multiple clusters for the same object, are resolved using non-maximum suppression (NMS). Additionally, ImmortalTracker \cite{wang2021immortal} performs multi-object tracking to filter noisy annotations and estimate novel object velocities.

\smallskipcustom

\noindent \textbf{Selective Alignment.} Our method motivates the idea of having only high-quality alignment pairs contribute to the alignment loss during training, while all 3D annotations remain included in the box-related losses. Prior works~\cite{leonardis_find_2025, cao_coda_2023} restrict alignment filtering to multi-view alignment methods, as formulated in Eq.~\eqref{eq:mv_align}. However, as illustrated in \cref{fig:3}, such filtering is insufficient in cases of partial occlusion or low resolution, where multi-view alignment appears nearly perfect. To address this, \NOD deploys additional filters to detect and exclude unreliable alignments, ensuring they do not contribute to the alignment loss during training.

 To mitigate \emph{occlusion-induced corruption}, the instance mask \( m \) of each novel object proposal is used to filter out pairs with inadequate object representation. Specifically, the proportion of pixels occupied by the object within the 2D crop is thresholded by \( \tau_\text{occ} \):
\begin{align}
\frac{\sum_{x=1}^{W_{\text{crop}}} \sum_{y=1}^{H_{\text{crop}}} m(x, y)}{H_{\text{crop}} W_{\text{crop}}}  \leq \tau_{\text{occ}}.
\end{align}
This process identifies and excludes occluded objects from cross-modal alignment. Different values of $\tau_{\text{occ}}$ are selected depending on the novel class, oultined in [supplementary]

Next, \NOD filters out samples associated with image crops having an insufficient number of pixels, as these are prone to \emph{resolution-induced} noise. Samples with a resolution below the threshold \(\tau_{\text{res}}\) are marked as having insufficient semantic resolution for discriminative 2D features:
\begin{align}
H_{\text{crop}} W_{\text{crop}} \leq \tau_{\text{res}}.
\end{align}

These strategies lead to more robust and accurate cross-modal alignment.

\subsection{Model Architecture} \label{sec:model-architecture}

\noindent \textbf{Overview.} The architecture of \ours comprises the off-the-shelf TransFusion-L~\cite{bai_transfusion_2022} 3D detector and our Hierarchal Two-Stage Alignment (H2SA) head. TransFusion-L processes a LiDAR point cloud to generate 3D object embeddings for novel objects and decodes their bounding boxes via its regression head. Meanwhile, H2SA transforms the 3D object embeddings into predicted alignment features for prompt-based classification.

\smallskipcustom

\begin{figure}
    \centering
    \includegraphics[width=1\linewidth]{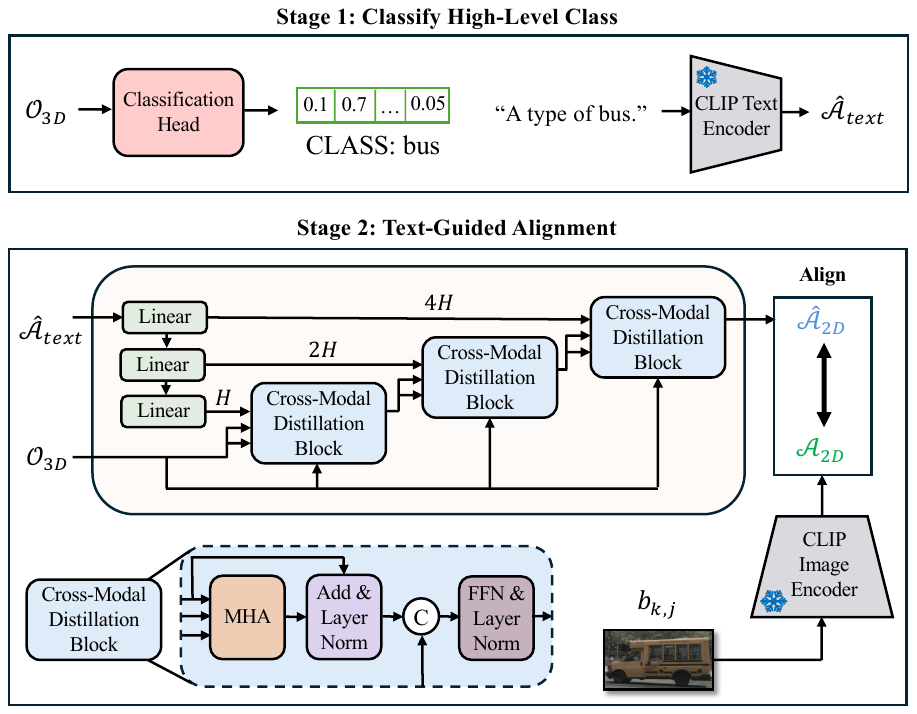}
    \caption{\textbf{Illustration of the Hierarchical Two-Stage Alignment (H2SA) Head.} H2SA first predicts the high-level novel classes, then derives class-based text prompts. H2SA then uses text prototypes to incrementally map 3D features to their 2D counterparts.}
    \label{fig:4}
\end{figure}

\noindent \textbf{Hierarchical Two-Stage Alignment (H2SA) Head.}
\label{sec:H2SA}
To enable strong semantic alignment, H2SA employs a hierarchical two-stage alignment process. Transfusion-L first predicts the high-level novel class \( c \) as an auxiliary task. A high-level text prompt \( \mathcal{\hat{A}}_{\text{text}} \in \mathbb{R}^{D}\) derived from the predicted \( c \) is then used to guide the alignment between 3D object embeddings \( \mathcal{O}_{\text{3D}} \in \mathbb{R}^{H} \) and target 2D image embeddings \( \mathcal{A}_{\text{2D}} \in \mathbb{R}^{D} \). The target embeddings are generated using the CLIP ViT-H/14 model, trained on DFN-5B~\cite{fang_data_2024}, which has strong zero-shot performance on ImageNet~\cite{deng_imagenet_2009} even at low resolutions. Since CLIP's image and text encoders produce high-dimensional embeddings with \( 8H = D \), H2SA upscales the 3D object embeddings to match this dimension by employing three embedding scales \( s \in \{ H, 2H, 4H \} \), as illustrated in \cref{fig:4}. Each scale transcribes the object's semantics into a scale-specific text prototype, progressively incorporating details for enhanced fine-grained alignment. Lastly, we note that our lightweight alignment head seamlessly integrates with TransFusion-L without compromising its deployability or real-time inference performance.

\smallskipcustom

\noindent \textbf{Multi-Sensor Fusion.} To show that \ours is also effective in the multi-sensor setting, we introduce \ours-Fusion, which integrates an image backbone following BEVFusion~\cite{liu_bevfusion_2023}.  We adopt pre-trained InternImage-T~\cite{wang2022internimage} to prevent distribution bias, since BEVFusion’s SWIN-T~\cite{liu2021swintransformerhierarchicalvision} backbone is pre-trained on NuImages~\cite{caesar2020nuscenes}.

\subsection{Training} \label{sec:training}
To train \ours, our method employs the TransFusion-L loss \cite{bai_transfusion_2022}. We incorporate an additional cross-modal alignment loss term into the overall loss function, where $S_c$ is cosine similarity.
\begin{align}
\ell_{\text{align}} &= \frac{\sum_{i=1}^{N_{\text{pos}}} \left(1 - S_{C}\left(\text{H2SA}(\mathcal{O}_\text{3D}^{(i)}, \mathcal{\hat{A}}_{\text{text}}^{(i)}), \mathcal{A}_\text{2D}^{(i)}\right)\right) \cdot \mathds{1}_{\text{align},i}}{\sum_{i=1}^{N_{\text{pos}}} \mathds{1}_{\text{align},i}}
\end{align}

All $N_{\text{pos}}$ positively matched prediction-target pairs contribute to each loss term, except for the alignment loss. Only pairs with targets selected by the selective alignment filter, \(\mathds{1}_{\text{align},i}\), contribute to the alignment loss. We adopt the same class-based data augmentation strategies as prior LiDAR-only works~\cite{zhu_class-balanced_2019}  by using novel class \(c\). Specifically, we balance each training sample by keeping a database of object point clouds and ensuring class distribution balance through the targeted injection of objects during training. 



\subsection{Prompt-based Classification}\
\label{sec:prompting}
\noindent To classify objects, \ours employs a prompt-based classification strategy by computing similarities between predicted alignment embeddings and text embeddings of fine-grained subclasses. Using this approach, high-level novel classes can be inferred by reverse mappings from fine-grained subclasses (e.g., a “minivan” is mapped to “car”). The fine-grained sub-class to high-novel class mapping are outlined in \cref{tab:specific_to_broad} from the supplementary material.

\begin{table*}[t!]
    \centering
    \small 
    \caption{\textbf{Results on nuScenes.} We report the overall mAP, NDS, and individual class APs. All classes are novel (\ie, no human-annotations are used to train except in CoDA). CoDA uses ``car" as a base-class. 
    Best in \textbf{bold} and ``*" denotes omitted methods for fair comparison.}
    \resizebox{\textwidth}{!}{%
    \begin{tabular}{@{}l|cc|cccccccccc}
        \toprule
        \textbf{Method} & \textbf{mAP} & \textbf{NDS} & \textbf{Car} & \textbf{Ped.} & \textbf{Truck} & \textbf{Motorcyc.} & \textbf{Bicyc.} & \textbf{T. Cone} & \textbf{Bus} & \textbf{Barrier} & \textbf{Con.V.} & \textbf{Trailer} \\
        \midrule
        OV-3DET~\cite{lu_open-vocabulary_2023} & 5.7  & 12.0 & 16.1 & 21.1 & 2.9  & 2.0  & 1.8  & 11.0 & 1.3  & 0.3  & 0.3  & 0.0 \\
        CoDA~\cite{cao_coda_2023} & 10.3  & 16.3 & 85.5 & 3.5 & 1.9  & 2.0  & 5.9  & 3.9 & 0.0  & 0.0  & 0.0  & 0.0 \\
        UP-VL~\cite{upvl_2024} & 11.3 & 17.2 & 19.4 & 30.8 & 7.9  & 16.8 & 14.1 & 15.6 & 3.1  & 0.5  & 4.3  & 0.4 \\
        Find n' Prop.~\cite{leonardis_find_2025} & 16.7 & 22.4 & 24.3 & 22.8 & 8.6  & 35.8 & \textbf{34.7} & 21.3 & 11.1 & 4.4  & 4.1  & 0.1 \\
        OpenSight~\cite{zhang_opensight_2024} & 22.9 & 23.3 & 25.3 & 52.5 & 11.6 & 30.4 & 26.1 & 25.6 & 5.1  & \textbf{42.2} & \textbf{8.7}  & \textbf{0.8} \\
        \midrule
        \rowcolor{gray!10}
        \ours & \textbf{31.1} & \textbf{32.8} & \textbf{61.6} & \textbf{60.1} & \textbf{30.3} & \textbf{39.8} & 31.0 & \textbf{39.6} & \textbf{22.0} & 18.8 & 6.8  & 0.6 \\
        \ours-Fusion* & 33.8 & 34.4 & 62.0 & 57.6 & 34.2 & 44.7 & 40.2 & 44.6 & 24.4 & 18.9 & 10.8  & 0.8 \\
        Supervised w/ Boxes* & 49.8 & 60.8 & 81.4 & 75.4 & 34.7  & 55.1 & 39.4 & 62.5 & 52.7  & 70.8 & 13.6  & 12.4 \\
        \bottomrule
    \end{tabular}%
    }
    \label{table1}
    \vspace{-1mm}
\end{table*}
\begin{table}[t!]
    \centering
    \small 
    \caption{\textbf{Results on KITTI.} We report the overall AP$_{3D@50}$ for each class at medium difficulty. All classes are novel.}
    \begin{tabular}{lccc|c}
        \toprule
        \textbf{Method} & \textbf{Car} & \textbf{Ped.} & \textbf{Cyclist} & \textbf{mAP$_{3D@50}$} \\
        \midrule
        OV-3DET~\cite{lu_open-vocabulary_2023} & 42.65 & 15.71 & 18.20 & 25.52 \\
        ImOV3D~\cite{yang2024imov3d} & \textbf{45.51} & 19.53 & 28.00 & 31.01 \\
        \midrule
        \rowcolor{gray!10}
        OV-SCAN & 45.10 & \textbf{27.61} & \textbf{29.81} & \textbf{34.17} \\
        \bottomrule
    \end{tabular}
    \label{table2}
    \vspace{-1mm}
\end{table}
\begin{table}[t!]
    \centering
    \small 
    \caption{\textbf{Ablations on Adaptive 3D Box Search.} We evaluate performance on nuScenes using different box search methods and varying numbers of search iterations per proposal.}
    \begin{tabular}{cc|c|ccc}
        \toprule
        \textbf{Method} & \textbf{Iter.} & \textbf{mAP} & \textbf{mAP$_{n}$} & \textbf{mAP$_{m}$} & \textbf{mAP$_{f}$} \\
        \midrule
        Greedy & 150,000 & 25.4 & 38.3 & 29.2 & 22.4 \\
        Adaptive & 37,500 & 28.9 & 47.8 & 33.8 & 25.4 \\
        Adaptive & 75,000 & 29.2 & 47.5 & 34.3 & 25.8 \\
        \midrule
        \rowcolor{gray!10}
        Adaptive & 150,000 & 31.1 & 48.1 & 35.6 & 27.7 \\
        \bottomrule
    \end{tabular}

    \label{table3}
\end{table}


\begin{table}[t!]
    \centering
    \small
    \caption{\textbf{Ablation Studies.} Selected settings are marked in \colorbox{gray!10}{gray}.}
    \begin{tabular}{cc|cc}
        \toprule
        \multicolumn{2}{c|}{\phantomsection\label{tab:ablation-cost-weights} (a) 3D Box Search Cost Weights} & \multicolumn{2}{c}{\phantomsection\label{tab:ablation-resolution-filter} (b) Resolution Filter} \\
        \rule{0pt}{10pt}
        $\boldsymbol{(\lambda_1, \lambda_2, \lambda_3, \gamma)}$ & \textbf{mAP} & \textbf{Filter} & \textbf{mAP} \\
        \midrule
         $(5.0, 0.0, 0.0, 3.0)$ & 26.1 & w/o & 30.4 \\
         $(1.0, 1.0, 1.0, 1.0)$ & 27.4 & $\tau_{\text{res}}=1000$ & 30.8 \\
         $(5.0, 2.5, 2.5, 3.0)$ & 27.8 & \cellcolor{gray!10}$\tau_{\text{res}}=4000$ & \cellcolor{gray!10}31.1 \\
         \cellcolor{gray!10}$(5.0, 1.0, 1.0, 3.0)$ & \cellcolor{gray!10}31.1 & $\tau_{\text{res}}=10000$ & 30.6 \\
        \bottomrule
        \toprule
        \multicolumn{2}{c|}{\phantomsection\label{tab:ablation-occlusion-filter} (c) Occlusion Filter} & \multicolumn{2}{c}{\phantomsection\label{tab:ablation-alignment-head} (d) H2SA Head} \\
        \rule{0pt}{10pt}
        \textbf{Filter} & \textbf{mAP} & \textbf{Alignment Head} & \textbf{mAP} \\
        \midrule
        w/o & 29.4 & One-Stage + FFN & 25.2 \\
        fixed ($\tau_{\text{occ}} = 0.35$) & 30.7 & Two-Stage + FFN & 29.4 \\
        \cellcolor{gray!10}class-based $\tau_{\text{occ}}$ & \cellcolor{gray!10}31.1 & \cellcolor{gray!10}H2SA & \cellcolor{gray!10}31.1 \\
        \bottomrule
    \end{tabular}
    \label{table:ablation-studies}
\end{table}

\section {Experiments} 

\subsection {Experimental Setup} 
\noindent  \textbf{Datasets.}\textbf{}
Our OV-3D object detection experiments are conducted on the nuScenes~\cite{caesar2020nuscenes} and KITTI~\cite{geiger2013kitti} datasets. We consider all classes as novel and utilize human-annotated labels solely for evaluation. 

\smallskipcustom

\noindent \textbf{Novel Object Discovery Setup.} In our experiments, novel classes (open-set classes) are based on the nuScenes and KITTI class labels. We use particle swarm optimization~\cite{kennedy_pso_1995} for \NOD's adaptive 3D box search, balancing the cost function with  \((\lambda_1, \lambda_2, \lambda_3, \gamma) = (5.0, 1.0, 1.0, 3.0)\)  and performing 150,000 iterations per cross-modal proposal. For more details on the implementation, please refer to \cref{sec: a3dbs} in the supplementary material.

\smallskipcustom

\noindent \textbf{Training Setup.} \ours is trained on 8 NVIDIA V100 GPUs with a batch size of 4 for 20 epochs. \ours adopts the AdamW optimizer with a learning rate 0.001 and a weight decay of 0.01. The CLIP image and text encoders are kept frozen.
For \ours-Fusion, we add an image backbone on top of pre-trained \ours and fine-tune. \ours-Fusion is trained for 5 additional epochs using a cosine annealing schedule initialized at a learning rate of 0.0001. This work is built on OpenPCDet~\cite{openpcdet2020}, an open-source 3D object detection toolbox implemented using PyTorch. We plan to release our code upon publication.

\smallskipcustom

\noindent \textbf{Evaluation Metrics.} For nuScenes, we evaluate performance using mAP and NDS, and 10 class APs. For KITTI, we compute APs for 3 classes using a stricter 3D IoU at 0.5 matching threshold. Following OpenSight~\cite{zhang_opensight_2024}, our ablation studies also report mAP$_{n/m/f}$ at three ranges: near (0-18m), mid (0-34m), and far (0-54m), respectively.

\begin{figure*}[t!]
    \centering
    \includegraphics[width=\textwidth]{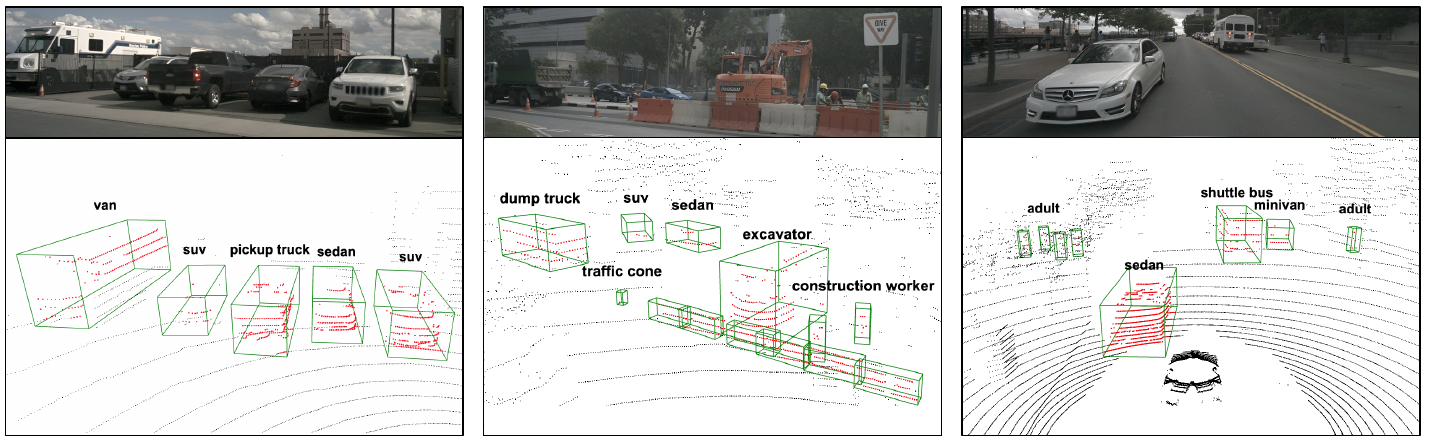} 
    \caption{\textbf{Visualization results using Open-Vocabulary.} \ours performs inference on a set of urban scenes identifying a diverse set of objects. To demonstrate OV capabilities, objects are classified into more fine-grained classes as opposed to traditional closed-set classes.
    }
    \label{fig:6}
\end{figure*}

\subsection{Main Results}
\label{sec:main_results}

\textbf{NuScenes.}  The OV-3D object detection results for top-performing methods on the nuScenes dataset can be seen in \cref{table1}. For a fair comparison to the other methods, which are all LiDAR-based, we only consider \ours and not \ours-Fusion. Our LiDAR-based method surpasses previous benchmarks by achieving the best mAP and NDS, thus surpassing OpenSight by 8.2 mAP and 9.5 NDS. Without being given 3D human-annotations, \ours achieves an AP score above 60 for both car and pedestrian categories. We also note that our method exceeds previous benchmarks by more than 5 AP for 6 out of the 10 classes. Furthermore, we show that simply adding camera as an additional input modality to \ours and then fine-tuning can improve the overall performance. Finally, as an upper bound, we report the performance of our method using ground truth boxes. CLIP is still used to derive high-level class labels and align.

\smallskipcustom

\noindent \textbf{KITTI.} In \cref{table2}, we present our results on KITTI, demonstrating the applicability of our method across multiple datasets. \ours outperforms OV-3DET~\cite{lu_open-vocabulary_2023} and ImOV3D~\cite{yang2024imov3d} in the overall metric, achieving comparable results to ImOV3D~\cite{yang2024imov3d} in the car category while surpassing both in the other two classes.

\begin{figure}[t!]
    \centering
    \begin{subfigure}[b]{0.4725\linewidth}
        \centering
        \includegraphics[width=\linewidth]{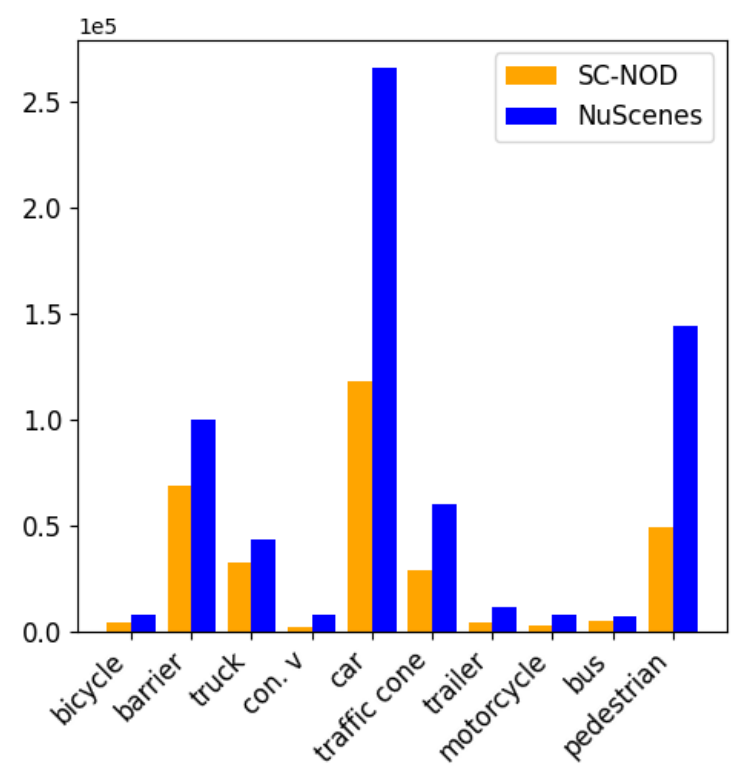}
        \caption{Annotation Count}
        \label{fig7a}
    \end{subfigure}
    \hfill
    \begin{subfigure}[b]{0.49\linewidth}
        \centering
        \includegraphics[width=\linewidth]{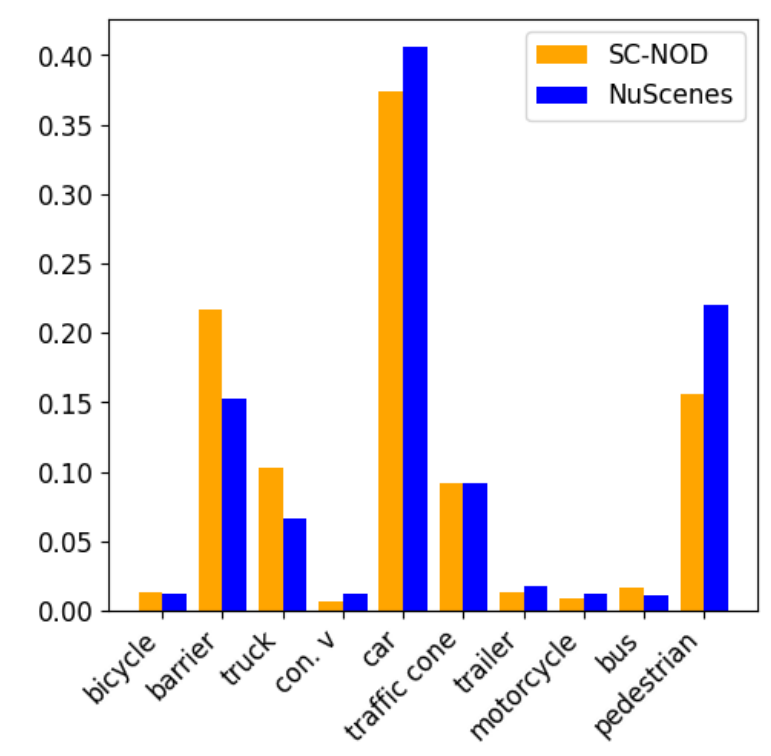}
        \caption{Relative Annotation Count}
        \label{fig7b}
    \end{subfigure}
    \caption{\textbf{Distribution of training labels.}}
    \label{fig:main}
\end{figure}

\noindent \textbf{Novel Object Discovery.} \NOD generates 319,028 3D annotations for training, a fraction of the 797,179 available in the nuScenes dataset. While all annotations contribute to the box loss, only 171,532 (54\%) of those generated are utilized for cross-modal alignment. The remainder of generated annotations are excluded as a result of filtering due to significant occlusion (39\%) or insufficient resolution (7\%). As shown in \cref{fig7b}, the relative annotation count of \NOD labels exhibits a comparable distribution.

\vspace{-1.5mm}

\subsection{Ablation Studies}

\noindent \textbf{Adaptive 3D Box Search.} To assess the effectiveness of our adaptive 3D box search in \NOD, we evaluate its performance on the nuScenes dataset, comparing it to the greedy search approach used in Find N’ Propagate~\cite{leonardis_find_2025}. \cref{table3} shows that our method consistently outperforms the greedy search even with fewer iterations per novel object. We further ablate various parameterizations of the cost function during the adaptive 3D box search, as seen in 
Tab. \protect\hyperref[tab:ablation-cost-weights]{\ref{tab:ablation-cost-weights}a}
, highlighting the importance of balancing the heuristic cost terms for effective 3D box search. 

\smallskipcustom

\noindent \textbf{Selective Alignment.} 
Tab. \protect\hyperref[tab:ablation-resolution-filter]{\ref{tab:ablation-resolution-filter}b}
and 
Tab. \protect\hyperref[tab:ablation-occlusion-filter]{\ref{tab:ablation-occlusion-filter}c} demonstrates the effectiveness of filtering techniques used in selective cross-modal alignment. A simple occlusion filter with a fixed threshold $\tau_{occ}$ yields a notable performance gain, while class-based thresholds achieve the highest improvement (+1.7 mAP). For the resolution filter, optimizing the threshold to balance resolution quality and the number of filtered samples proves most effective (+0.7 mAP).

\smallskipcustom

\noindent \textbf{Alignment Head.} To assess the impact of the H2SA head, we introduce a one-step baseline in 
Tab. \protect\hyperref[tab:ablation-alignment-head]{\ref{tab:ablation-alignment-head}d}
for comparison. This variant removes the classification loss term, merges TransFusion-L’s class heatmaps into a single class-agnostic heatmap, and replaces the text-guided alignment network with a simple feed-forward network. This simplification results in a performance drop (-5.9 mAP). Furthermore, incorporating class-based text prompts to guide cross-modal alignment further enhances performance (+1.7 mAP). 

\subsection{Limitations} 
The primary limitation of SC-NOD is its limited annotation recovery (\cref{fig7a}), due to reliance on 2D proposals. Objects clearly visible in the point cloud may remain unlabeled if absent in multi-view images, resulting in unlabeled objects. Additionally, labeling sparse objects at long distances remains challenging, significantly reducing performance at greater ranges (\cref{table3}). Furthermore, for 3D box search, \NOD adopts class-based anchors following prior works. While suitable for novel classes with low dimensional variance, this method struggles with highly shape-variable objects such as trailers, buses, and construction vehicles. These insights motivate future work exploring alternative methods less dependent on 2D proposals and anchor-free box-parameterization strategies.

\section{Conclusion}

We introduce OV-SCAN, an OV-3D detector that achieve detection through enhanced cross-modal alignment. Without human-provided 3D annotations, SC-NOD accurately generates 3D boxes and also carefully guides cross-modal alignment. By adapting H2SA, we further strengthen alignment and enable robust open-set classification. Experiments on nuScenes demonstrate state-of-the-art OV-3D detection.

{
    \small
    \bibliographystyle{ieeenat_fullname}
    \bibliography{main}
}
\clearpage
\setcounter{page}{1}
\maketitlesupplementary

\section{Additional Implementation Details}
\label{sec:details}

\subsection{Novel Object Proposals}

\ours relies on Grounding DINO~\cite{liu2024DINO} and SAM~\cite{kirillov_segment_2023} to detect 2D proposals from a given list of (user-defined) novel classes. \cref{tab:Grounding Dino Text Prompts} outlines the novel classes we used for our experiments for both the nuScenes~\cite{caesar2020nuscenes} and KITTI~\cite{geiger2013kitti} datasets. Notably, we adhere to each dataset’s original taxonomy, with the exception of splitting “cyclist” into “bicycle” and “motorcycle” for KITTI.

\begin{table}[h!]
\centering
\small
\caption{\textbf{Novel classes used to identify novel object proposals for each dataset.} These classes are referred to as ``novel'' since their ground truth labels, although available in the respective datasets, are not used in training \ours.}
\begin{tabular}{cp{0.35\textwidth}} 
\toprule
\textbf{Dataset} & \textbf{Novel Classes} \\ 
\midrule
\centering nuScenes & car, truck, pedestrian, bicycle, motorcycle, bus, traffic cone, barrier block, construction vehicle  \\ 
\centering KITTI & car, van, truck, tram, bicycle, motorcycle, pedestrian, person sitting  \\
\bottomrule
\end{tabular}

\label{tab:Grounding Dino Text Prompts}
\end{table}
In post-processing, our method adopts a similar approach to UP-VL~\cite{upvl_2024} for removing false positives. We treat each novel class as a positive class while including a set of background classes. To refine the results, we cross-reference each image crop from Grounding DINO with CLIP~\cite{radford2021CLIP}, filtering out background classes such as “vegetation”, “fence”, “gate”, “curb”, “sidewalk”, “wall”, “building”, “railing”, and “rail guard.”

\subsection{Cross-Modal Association}

As discussed in~\cref{sec:scnod}, the primary objective of cross-modal association is to accurately pair each 2D proposal with its corresponding 3D object cluster. Our detailed implementation is outlined in~\cref{alg:cross_modal_asso}, where each 2D proposal is first projected into a 3D frustum defined by $(d_{min}, d_{max})$. Ideally, each proposal can then be matched to the cluster containing the point closest to the center frustum ray, provided this distance is within a matching threshold $\tau_{match}$. However, challenges arise when a single 2D proposal corresponds to multiple object clusters due to fragmentation from misclustering, partial occlusion, or sparsity. To address this, we allow \textit{one-to-many} matching, enabling a single 2D proposal to generate multiple competing cross-modal proposals. We resolve such conflicts during cross-modal target preparation by optimizing 3D box parameters for each candidate proposal, ultimately retaining the optimal proposal based on the objective defined in~\cref{eq:opt}. Additionally, we address scenarios involving \textit{many-to-one} matching, where larger objects (e.g., transit buses) span multiple camera views. In these situations, applying 3D NMS at the conclusion of cross-modal target preparation effectively resolves potential redundancies. Our conflict resolution strategy is detailed in~\cref{alg:cross_modal_target_prep}.

\label{sec:cma_supp}

    \begin{algorithm}
    \caption{\textbf{Cross-Modal Association}}
    \label{alg:cross_modal_asso}
    \begin{lstlisting}[language=Python, basicstyle=\ttfamily\scriptsize, 
                       keywordstyle=\bfseries, commentstyle=\color{darkgreen}]
def cross_modal_association(3D_object_clusters, 
                            2D_novel_object_proposals, 
                            calibrations):
    """
    Input:
    - 3D_object_clusters: Set of 3D object clusters
    - 2D_novel_object_proposals: Set of 2D novel
                                 object proposals
    - calibrations: Intrinsic and extrinsic 
                    calibration parameters
    
    Output:
    - cross_modal_proposals: 
            List of (2D proposal, 3D cluster) pairs
    """

    # Initialize proposal list
    cross_modal_proposals = []  

    for proposal in 2D_novel_object_proposals:
        # Compute frustum from 2D proposal
        frustum = get_frustum(proposal, calibration, 
                              d_min, d_max)
        
        # Compute frustum center ray
        frustum_center_ray 
            = get_frustum_center_ray(frustum)
        
        # Find 3D clusters with tau distance 
        # from center frustum ray
        matched_clusters 
            = get_clusters_near_ray(3D_object_clusters,
                                    frustum_center_ray,
                                    tau_match)
        
        # Store matching pairs 
        for cluster in matched_clusters:
            pair = (proposal, cluster)
            cross_modal_proposals.append(pair)

    return cross_modal_proposals
    \end{lstlisting}
    \end{algorithm}

    \begin{algorithm}
    \caption{\textbf{Cross-Modal Target Preparation}}
    \label{alg:cross_modal_target_prep}
    \label{alg:cross_modal}
    \begin{lstlisting}[language=Python, basicstyle=\ttfamily\scriptsize, 
                       keywordstyle=\bfseries, commentstyle=\color{darkgreen}]
def cross_modal_target_preparation(cross_modal_proposals, 
                                   novel_object_bank,
                                   calibrations):
    """
    Input:
    - cross_modal_proposals: 
        List of (2D proposal, 3D cluster) pairs
    - novel_object_bank: 
        Set of novel objects for training 
    - calibrations: Intrinsic and extrinsic 
                    calibration parameters
    """

    cross_modal_targets = []
    box_search_cost_costs = []

    for pair in cross_modal_proposals:

        2d_proposal, 3d_object_cluster = pair
    
        # Fit a 3D bounding box to the proposal.
        # 3D_box_params = (x,y,z,l,w,h,ry)
        3D_box_params, box_search_cost = 
            adaptive_3D_box_search(3d_object_cluster,
                                   2d_proposal, 
                                   PSO_params)
        
        # Get instance mask (from SAM)
        instance_mask = get_instance_mask(2d_proposal) 

        # Selective Alignment Filters
        is_not_occluded = occlusion_filter(2d_proposal,
                                       instance_mask)
        is_high_res = resolution_filter(2d_proposal)
        is_aligned_mv = 
            multi_view_alignment_filter(2d_proposal,
                                        3D_box_params,
                                        calibrations)
        fit_for_alignment =
          is_not_occluded & is_high_res & is_aligned_mv
                                                    
        # Get 2D Embedding from CLIP
        2D_image_embed = CLIP(2d_proposal)
        high_level_novel_class = classify(
            2D_image_embed,
            set_of_novel_classes
        )

        # Prepare novel object target
        novel_object_target = (
            3D_box_params,               
            2D_image_embed,
            high_level_novel_class,
            fit_for_alignment
        )
        
        cross_modal_targets.append(novel_object_target)
        box_search_costs.append(box_search_cost)

    # Solve conflicts from one-to-many matching in CMA
    cross_modal_targets = resolve_CMA_conflicts(
        cross_modal_targets, 
        box_search_costs
    )

    # Perform NMS to remove duplicates
    cross_modal_targets =
        NMS(cross_modal_targets, box_search_costs)

    # Update novel object bank
    update(novel_object_bank, cross_modal_targets) 
    
    return
    \end{lstlisting}
    \end{algorithm}

\subsection{Adaptive 3D Box Search}
\label{sec: a3dbs}

To solve the continuous nonlinear optimization problem from \cref{eq:opt}, \ours employs particle swarm optimization (PSO)~\cite{kennedy_pso_1995} to search for the 3D annotation parameterized by \( \theta = (x, y, z, l, w, h, r_y)\). Each object proposal corresponds to a unique optimization problem. The control hyperparameters for PSO are reported in \cref{tab:pso_parameters}.

\begin{table}[h!]
\centering
\small
\caption{PSO hyperparameters used in adaptive 3D box search.}
\begin{tabular}{ccc}
\toprule
\textbf{Parameter} & \textbf{Description} & \textbf{Value} \\ \midrule
\(N_{\text{swarm}}\) & Swarm size & 50 \\ 
\(N_{\text{iter}}\) & Iterations per particle & 3000 \\ 
\(w_{\text{init}}\) & Initial inertia weight & 10.0 \\ 
\(w_{\text{end}}\) & End inertia weight & 0.1 \\ 
\(c_1\) & Cognitive coefficient & 1.0 \\ 
\(c_2\) & Social coefficient & 1.0 \\ 
\(C_{\text{noise}}\) & Initialization noise & 0.1 \\ 
\bottomrule
\end{tabular}
\label{tab:pso_parameters}
\end{table}

For each PSO search, the position values \((x, y, z)\) of each candidate are initialized in 
areas with a high likelihood of corresponding to the true object center. In particular, half of the candidates are initialized at the closest point to the center frustum ray, and the rest are initialized at the mean of the object point cluster \(\mathcal{P}_{\text{obj}}\). To accommodate larger objects that require a broader search space, noise proportional to the anchor's size is sampled from \(\mathcal{N}(0, C_{\text{noise}} (\frac{\mathbf{A}_{\text{max}}+\mathbf{A}_{\text{min}}}{2}))\) and added to the initialized position. The dimension and orientation parameters are initialized uniformly across candidates. The inertia weight \(w\) follows a cosine annealing schedule to balance exploration with exploitation. We follow Find n' Propagate's ~\cite{leonardis_find_2025} class anchors.

\subsection{Selective Alignment} 

\begin{figure}[h!]
    \centering
    \includegraphics[width=1\linewidth]{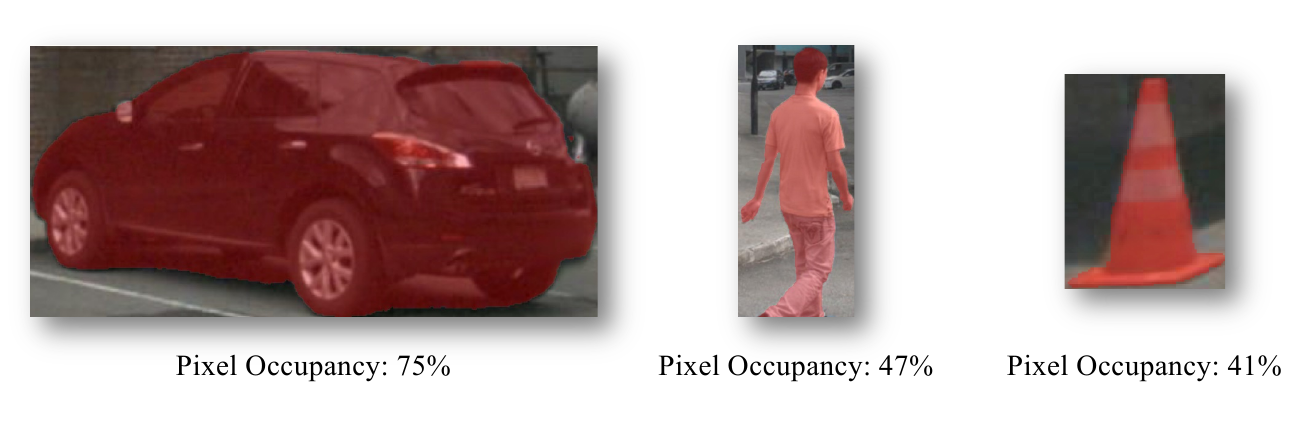}
    \caption{\textbf{Percentage of instance pixels across object classes.} Vehicles, such as cars, typically occupy a greater proportion of pixels within their 2D proposals compared to objects like pedestrians and traffic cones.}
    \label{fig:occupancy}
\end{figure}

During selective alignment, \ours uses class-specific thresholds $\tau_{occ}$ for the occlusion fileter since objects naturally occupy varying amounts of space within a 2D proposal. As illustrated in~\cref{fig:occupancy}, a car generally occupies a larger area in its instance mask compared to a pedestrian, resulting in a significantly higher proportion of pixels classified as instance pixels. Consequently, the pixel distribution within instance masks varies considerably across classes, making a uniform threshold for determining occlusion level inadequate.

To address this variability, reasonable threshold values for $\boldsymbol{\tau_{occ}}$ are manually determined for each class, as shown in~\cref{tab:taU_occ_params}, ensuring more robust estimation of highly occluded objects.

\begin{table}[h!]
\centering
\small
\caption{Parameterization of $\tau_{occ}$ for different novel classes.}
\label{tab:taU_occ_params}
\begin{tabular}{cc}
\toprule
\textbf{Novel Class} & $\boldsymbol{\tau_{occ}}$ \\ 
\midrule
car & 0.5 \\
truck & 0.5 \\
pedestrian & 0.25 \\
bicycle & 0.4 \\
motorcycle & 0.4 \\
bus & 0.5 \\
traffic cone & 0.25 \\
barrier block & 0.35 \\
construction vehicle & 0.5 \\
\bottomrule
\end{tabular}
\end{table}

\subsection{Hierarchical Two-Stage Alignment Head} 

In Stage 1, H2SA employs classification as an auxiliary task to generate high-level text prompts for alignment. Following TransFusion-L~\cite{bai_transfusion_2022}, it regresses class-specific heatmaps to jointly localize and classify object proposals. We compute the class-based text prompt embeddings $\hat{A}_{\text{text}}$ ahead of time for retrieval. The top \textit{K} proposals are then passed through an object decoder to produce the set of features $mathcal{O}_{\text{3D}}$. In stage two, H2SA aligns each 3D object embedding \(\mathcal{O}_{\text{3D}}\) with its 2D counterpart \(\mathcal{A}_{\text{2D}}\). H2SA passes $\hat{A}_{\text{text}}$ through a set of linear layers to generate multi-scale text prototypes \(\{W_{H}, W_{2H}, W_{4H}\}\). The Cross-Modal Distillation Block (CMDB) refines and upscales these prototypes, distilling \(\mathcal{O}_{\text{3D}}\) into multi-scale representations. The first-step CMDB operation is defined as:
\begin{align}
W_{H}' = \text{LN}(\text{MHA}(W_{H}, \mathcal{O}_{\text{3D}}, \mathcal{O}_{\text{3D}}) + W_{H}),
\end{align}
where \(\text{LN}\) is layer normalization and \(\text{MHA}\) is multi-head attention. Next, CMDB fuses the refined text prototype \(W_{H}'\) with the 3D object embedding \(\mathcal{O}_{\text{3D}}\) to produce a unified feature \(U_{2H}\) for the next step. This fusion is achieved through channel-wise concatenation, followed by a feed-forward network:
\begin{align}
U_{2H} = \text{LN}(\text{FFN}(\text{concat}(W_{H}', \mathcal{O}_{\text{3D}}))).
\end{align}

This process iteratively integrates features across different scales, enabling robust cross-modal alignment to higher-dimension alignment targets.
 
\subsection{Prompt-based Classification}  

As mentioned in \cref{sec:prompting}, \ours employs a specific-to-broad strategy, relying exclusively on H2SA for classification. First objects are classified into fine-grained subclasses before being mapped to their respective novel classes for evaluation. To achieve this, we utilize the frozen CLIP text encoder to generate text embeddings using the template \texttt{"a type of \{SUBCLASS\}."} as the text prompt. For each object proposal generated by the detector, the fine-grained subclass is determined by selecting the subclass with the highest object-text similarity. The corresponding label \(\hat{c}_{\text{fg}}\) for each object proposal is computed as:
\begin{align}
\hat{c}_{\text{fg}} = \arg\max_{c_i \in C_{\text{fg}}} S_C(\text{H2SA}({\mathcal{O}}_{\text{3D}}, \mathcal{\hat{A}}_{\text{text}}), t_{c_i}),
\end{align} where \(C_{\text{fg}}\) denotes the set of fine-grained subclasses and \(t_{c_i}\) is the text embedding corresponding to subclass \(c_i\). For the nuScenes dataset, the fine-grained subclasses used are outlined in \cref{tab:specific_to_broad}. \ours follows the same procedure for the KITTI dataset.

\begin{table}[h!]
\centering
\small
\renewcommand{\arraystretch}{1.2}
\label{tab:specific_to_broad}
\caption{\textbf{Fine-grained subclasses for each high-level novel class in the nuScenes dataset.} Con.V. refers to construction vehicle.}
\begin{tabular}{cp{0.33\textwidth}} 
\toprule
\textbf{Novel Class} & \textbf{Fine-grained Subclasses} \\ 
\midrule
car & sedan, van, minivan, hatchback, suv, coupe, police car, sprinter van, taxi \\ 

truck & pickup truck, tow truck, semi-truck, gasoline truck, delivery truck, garbage truck, fire truck, flatbed truck, ambulance, cement truck, dump truck \\ 

bus & school bus, coach bus, double-decker bus, transit bus, shuttle bus, minibus \\ 

trailer & portable message board trailer, flatbed trailer, freight trailer, cargo trailer \\ 

Con.V. & excavator, bulldozer, forklift, construction loader, construction lift \\ 

pedestrian & adult, construction worker, police officer, child \\ 

motorcycle & cruiser motorcycle, sport motorcycle, touring motorcycle, moped \\ 

bicycle & bicycle \\ 

traffic cone & traffic cone, traffic drum, traffic delineator post\\ 

barrier & plastic jersey barrier, concrete jersey barrier \\ 
\bottomrule
\end{tabular}
\end{table}

\section{Extending to Additional Novel Classes}
This section outlines steps to expand the open set of novel classes:

\begin{figure*}[t!]
    \centering
    \includegraphics[width=\textwidth]{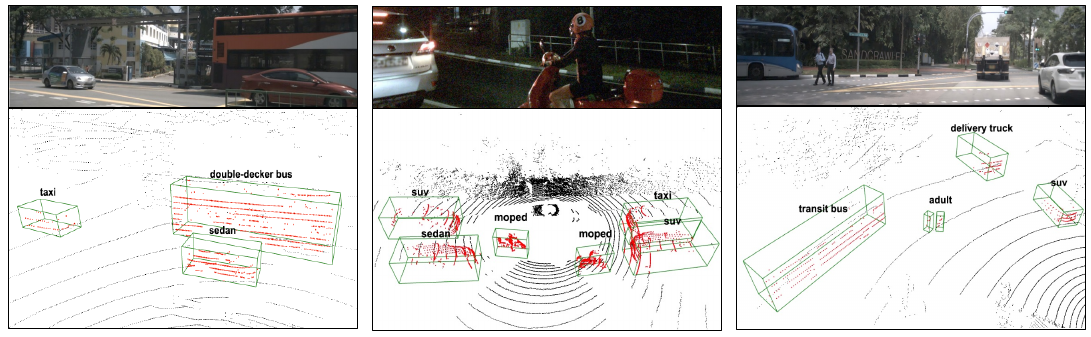} 
    \caption{\textbf{Visualization results on various traffic scenarios.} \ \ours detects novel objects in scenarios including oncoming traffic (left), vehicles stopped at a traffic light (center), and objects encountered at a busy intersection (right).
    }
    \label{fig:9}
\end{figure*}

\begin{enumerate}
    \item Extend the existing set of novel classes provided to Grounding DINO and SAM. Regenerate the set of novel object proposals for the given dataset.
    \item For each additional novel class, add additional anchor boxes for the adaptive 3D box search and regenerate the novel object bank for the dataset.
    \item Update the BEV encoder in TransFusion-L to incorporate heatmaps for the newly added high-level novel classes and train \ours.
    \item For prompt-based classification, provide the set fine-grained subclasses for each additional novel class.
\end{enumerate}

\begin{table}[t!]
    \centering
    \small
    \begin{minipage}{\linewidth}
        \centering
        \caption{\textbf{Results for various weather conditions on the nuScenes validation set.}}
        \begin{tabular}{c|c|c|c|c}
            \toprule
            \textbf{Method} & Day & Night & Dry & Wet \\
            \midrule
            OV-SCAN & 31.1 & 23.4 & 31.8 & 25.6 \\
            OV-SCAN-Fusion & 34.0 & 20.7 & 34.4 & 29.6 \\
            \bottomrule
        \end{tabular}
        \label{table:adv_weather}
    \end{minipage}
\end{table}

\section{Robustness of Multi-Sensor Fusion}

In \cref{table:adv_weather}, we observe that \ours-Fusion achieves improved detection performance over \ours across all weather conditions except at night. While the fusion strategy generally enhances robustness by leveraging complementary modalities, its effectiveness diminishes in low-light conditions. The reduced reliability of image features at night introduces noise into the model, highlighting a limitation in sensor fusion under varying lighting conditions.

\section{Additional Visualizations}

\textbf{Cross-Modal Alignment.} \cref{fig:9} presents additional qualitative results to highlight the cross-modal alignment performance. \ours detects two cars passing a parked double-decker bus in oncoming traffic. It also accurately identifies a pair of mopeds stopped at a traffic light. Finally, a variety of objects are detected at a busy intersection.

\smallskip

\begin{figure*}[h!]
    \centering
    \includegraphics[width=0.86\textwidth]{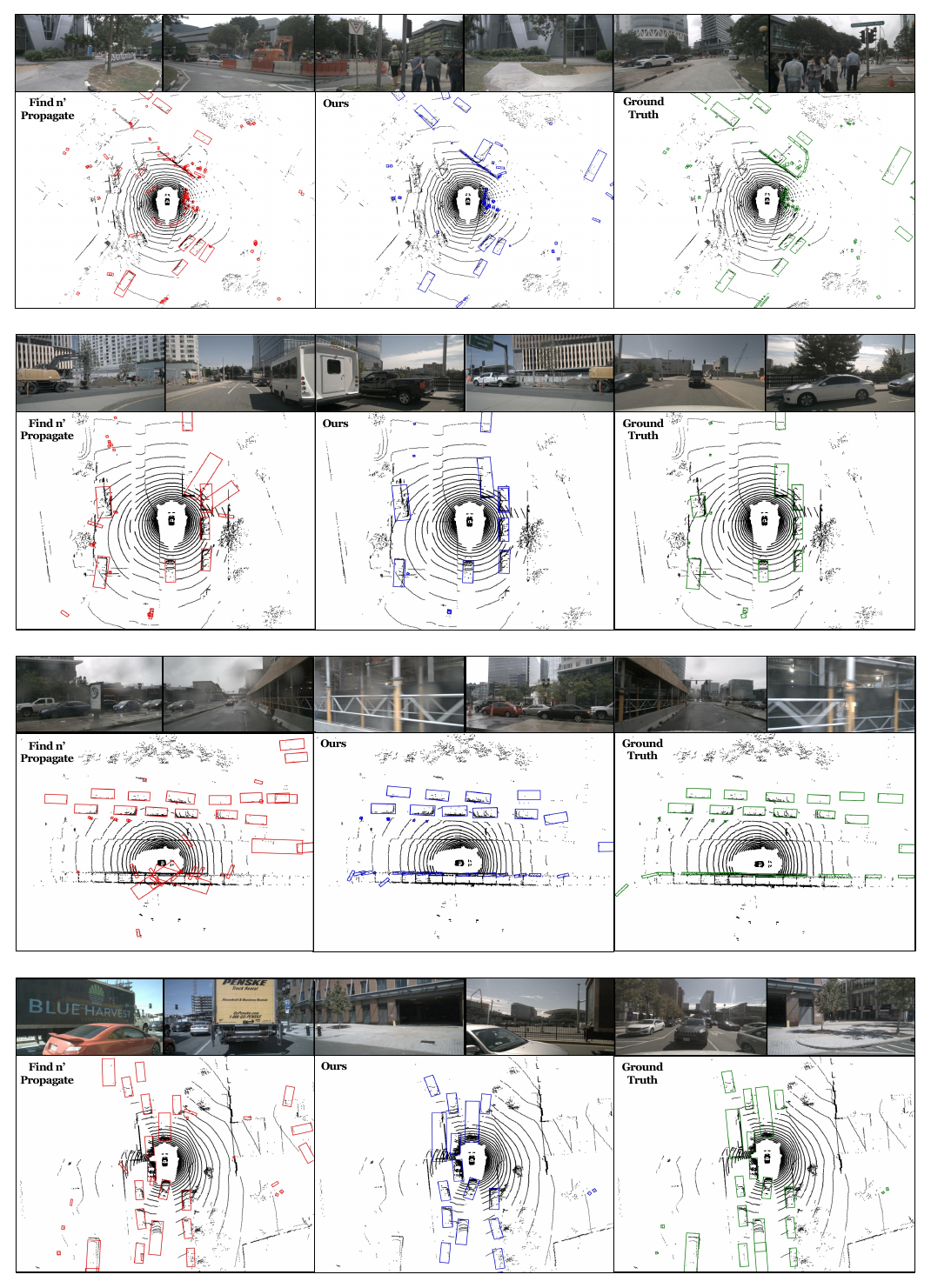} 
    \caption{\textbf{Comparison between \ours and Find n' Propagate~\cite{leonardis_find_2025}.} \ours regresses more precise bounding boxes than Find n' Propagate without requiring any human-annotated labels. We compare to Find n' Propagate in \textit{Setting 2} which uses 3 base classes ("car", "pedestrian", and "bicycle") leaving the rest as novel. 
    }
    \label{fig:10}
\end{figure*}

\noindent \textbf{3D Box Search.} In \cref{fig:10}, we perform a side-by-side comparison of the box regressed when fueled by \NOD vs. Greedy Box Seeker from Find n' Propagate. Both methods use Transfusion-L~\cite{bai_transfusion_2022} off-the-shelf, while displaying detections with a confidence over 0.05. However, during training, Find n' Propagate natively follows its predefined \textit{Setting 2}, treating three classes (“car,” “pedestrian,” and “bicycle”) as base and leaving the remaining classes as novel. In contrast, \ours treats every class as novel. \ours regresses notably more precise bounding boxes for novel classes. Additionally, Find n' Propagate tends to produce more false positives due to their increased recall strategy, whereas \ours maintains better precision and localization accuracy

\end{document}